\pdfoutput=1
\documentclass[11pt]{article}

\usepackage[utf8]{inputenc}

\usepackage{subfigure}
\usepackage{url}
\usepackage{graphicx}
\usepackage{amsmath,amssymb}
\usepackage{bm,bbm}
\usepackage{hyperref}
\usepackage{titlesec}
\titleformat{\section}
{\normalfont\large\bfseries\filcenter}{\thesection.}{1 ex}{}
\titleformat{\subsection}[runin]
{\normalfont\normalsize\bfseries\filcenter}{\thesubsection.}{1 ex}{}

\usepackage[charter]{mathdesign}
\usepackage[mathcal]{eucal}

\usepackage[margin=1.2 in]{geometry}

\newcommand{\eop}{\hfill$\square$}

\newtheorem{Satz}{Theorem}[section]
\newtheorem{Thm}[Satz]{Theorem}

\newtheorem{Prop}[Satz]{Proposition}

\newtheorem{Cor}[Satz]{Corollary}

\newtheorem{Lem}[Satz]{Lemma}

\newtheorem{Rem}[Satz]{Remark}
\newtheorem{Def}[Satz]{Definition}

\newtheorem{Ex}[Satz]{Example}
\newtheorem{Not}[Satz]{Notations}

\newtheorem{Ass}[Satz]{Assumption}

\newcommand{\Z}{\ensuremath{\mathbb{Z}}}        %

\newcommand{\diag}{\operatorname{diag}}
\newcommand{\sgn}{\operatorname{sgn}}
\newcommand{\abs}{\operatorname{abs}}

\newcommand{\Var}{\operatorname{Var}}

\newcommand{\calE}{\mathcal{E}}

\newcommand{\calP}{\mathcal{P}}

\makeatletter
\providecommand*{\diff}%
{\@ifnextchar^{\DIfF}{\DIfF^{}}}
\def\DIfF^#1{%
\mathop{\mathrm{\mathstrut d}}%
\nolimits^{#1}\gobblespace
}
\def\gobblespace{%
\futurelet\diffarg\opspace}
\def\opspace{%
\let\DiffSpace\!%
\ifx\diffarg(%
\let\DiffSpace\relax
\else
\ifx\diffarg\[%
\let\DiffSpace\relax
\else
\ifx\diffarg\{%
\let\DiffSpace\relax
\fi\fi\fi\DiffSpace}
\makeatother

\usepackage{graphicx} %
\usepackage{subfigure}

\usepackage[round]{natbib}

\usepackage{algorithm}
\usepackage{algorithmic}

\usepackage{hyperref}

\begin{document}
\title{	Obtaining Error-Minimizing Estimates and Universal Entry-Wise Error Bounds for Low-Rank Matrix Completion}
\author{Franz J. Király\thanks{Machine Learning Group, TU-Berlin, \url{franz.j.kiraly@tu-berlin.de}} \and
Louis Theran \thanks{Discrete Geometry Group, FU Berlin, \url{theran@math.fu-berlin.de}}
}
\date{}
\maketitle

\begin{abstract}
We propose a general framework for reconstructing and denoising single entries of incomplete and noisy entries.
We describe: effective algorithms for deciding if and entry can be reconstructed and, if so, for reconstructing and
denoising it; and a priori bounds on the error of each entry, individually.  In the noiseless case our algorithm is exact.
For rank-one matrices, the new algorithm is fast, admits a highly-parallel implementation, and produces an error
minimizing estimate that is qualitatively close to our theoretical and the state-of-the-are Nuclear Norm and OptSpace methods.
\end{abstract}

\section{Introduction}

Matrix Completion is the task to reconstruct low-rank matrices from a subset of its entries and occurs
naturally in many practically relevant problems, such as missing feature imputation,
multi-task learning~\citep{ArgMicPonYin08}, transductive learning~\citep{GolZhuRecXuNow10},
or collaborative
filtering and link prediction~\citep{SreRenJaa05,AcaDunKol09,MenElk11}.

Almost all known methods performing matrix completion are optimization methods such as the max-norm and
nuclear norm heuristics~\citep{SreRenJaa05,CanRec09, THK10}, or OptSpace~\citep{KMO10}, to name a few amongst many.

These methods have in common that in general (a) they reconstruct the whole matrix and (b) error bounds are given
for all of the matrix, not single entries. These two properties of existing methods are in particular
unsatisfactory\footnote{While the existing methods may be applied to a submatrix, it is always at the cost of accuracy if the
data is sparse, and they do not yield statements on single entries.} in the scenario when one is interested only in
predicting (resp.~imputing) one single missing entry or a set of interesting missing entries instead of all - which is for
real data a more natural task than imputing all missing entries, in particular in the presence of large scale data (resp.~big data).

Indeed the design of such a method is not only desirable but also feasible, as the results of \cite{KTTU12} suggest by relating
algebraic combinatorial properties and the low-rank setting to the reconstructability of the data. Namely, the authors provide
algorithms which can decide for one entry if it can be - in principle - reconstructed or not, thus yielding a statement of
trustability for the output of any algorithm\footnote{The authors also provide an algorithm for reconstructing some missing
entries in the arbitrary rank case, but without obtaining global or entry-wise error bounds, or a strategy to reconstruct
all reconstructible entries.}.

In this paper, we demonstrate the first time how algebraic combinatorial techniques, combined with stochastic error minimization,
can be applied to (a) reconstruct single missing entries of a matrix and (b) provide lower variance bounds for the error of any
algorithm resp.~estimator for that particular entry - where the error bound can be obtained without actually reconstructing the
entry in question. In detail, our contributions include:
\begin{itemize}
\item the construction of a variance-minimal and unbiased estimator for any fixed missing entry of a rank-one-matrix, under the assumption of known noise variances
\item an explicit form for the variance of that estimator, being a lower bound for the variance of any unbiased estimation of any fixed missing entry and thus yielding a quantiative measure on the trustability of that entry reconstructed from any algorithm
\item the description of a strategy to generalize the above to any rank
\item comparison of the estimator with two state-of-the-art optimization algorithms (OptSpace and nuclear norm), and error assessment of the three matrix completion methods with the variance bound
\end{itemize}
Note that most of the methods and algorithms presented in this paper restrict to rank one.  This is not, however,
inherent in the overall scheme, which is general.  We depend on rank one only in the sense that we understand
the combinatorial-algebraic structure of rank-one-matrix completion exactly, whereas the behavior in higher
rank is not yet as well understood.  Nonetheless, it is, in principle accessible, and, once available will
can be ``plugged in'' to the results here without changing the complexity much.

\section{The Algebraic Combinatorics of Matrix Completion}\label{sec:theory}
\subsection{A review of known facts}
In~\cite{KTTU12}, an intricate connection between the algebraic combinatorial structure, asymptotics of graphs
and analytical reconstruction bounds has been exposed. We will refine some of the theoretical concepts presented
in that paper which will allow us to construct the entry-wise estimator.
\begin{Def}
An matrix $M\in \{0,1\}^{m\times n}$ is called mask. If $A$ is a partially known matrix, then the mask of $A$ is the mask which has $1$-s in exactly the positions which are known in $A$; and $0$-s otherwise.
\end{Def}

\begin{Def}
Let $M$ be an $(m\times n)$ mask. We will call the unique bipartite graph $G(M)$ which has $M$ as bipartite adjacency
matrix the {\it completion graph} of $M$. We will refer to the $m$ vertices of $G(M)$ corresponding to the rows of $M$
as blue vertices, and to the $n$ vertices of $G(M)$ corresponding to the columns as red vertices. If $e=(i,j)$ is an
edge in $K_{m,n}$ (where $K_{m,n}$ is the complete bipartite graph with $m$ blue and $n$ red vertices), we will
also write $A_e$ instead of $A_{ij}$ and for any $(m\times n)$ matrix $A$.
\end{Def}
A fundamental result, \citep[Theorem 2.3.5]{KTTU12}, says that identifiability and reconstructability
are, up to a null set, graph properties.
\begin{Thm}\label{Thm:closure}
Let $A$ be a generic\footnote{In particular, if $A$ is sampled from a continuous density, then the set of
non-generic $A$ is a null set.}
and partially known $(m\times n)$ matrix of rank $r$, let $M$ be the mask of $A$, let $i,j$ be integers.
Whether $A_{ij}$ is reconstructible (uniquely, or up to finite choice) depends only on $M$ and the true rank $r$;
in particular, it does not depend on the true $A$.
\end{Thm}
For rank one, as opposed to higher rank, the set of reconstructible entries is easily obtainable from $G(M)$ by combinatorial means:
\begin{Thm}[{\citep[Theorem 2.5.36 (i)]{KTTU12}}]\label{Thm:closure}
Let $G\subseteq K_{m,n}$ be the completion graph of a partially known $(m\times n)$ matrix $A$. Then the set of uniquely
reconstructible entries of $A$ is exactly the set $A_e$, with $e$ in the transitive closure of $G$.
In particular, all of $A$ is reconstructible if and only if $G$ is connected.
\end{Thm}

\subsection{Reconstruction on the transitive closure}
We extend Theorem~\ref{Thm:closure}'s theoretical reconstruction guarantee by describing an explicit,
algebraic algorithm for actually doing the reconstruction.  This algorithm will be the basis
of an entry-wise, variance-optimal estimator in the noisy case.
In any rank, such a reconstruction rule can be obtained by exposing equations which explicitly give
known and unknown entries in terms of only known entries  due to the fact that the set of low-rank matrices is an irreducible
variety  (the common vanishing locus of finitely many polynomial equations).  We are able to derive the reconstruction
equations for rank one.
\begin{Def}
Let $P\subseteq K_{m,n}$ (resp. $C\subseteq K_{m,n}$) be a path (resp. cycle), with a fixed start and end
(resp. traversal order).  We will denote by $E^+(P)$ be the set of edges in $P$ (resp. $E^+(C)$ and $C$) traversed
from blue vertex to a red one, and by $E^-(P)$ the set of edges traversed from a red vertex to a blue one
\footnote{This is equivalent to fixing the orientation of
$K_{m,n}$ that directs all edges from blue to red, and then taking $E^+(P)$ to be the set of edges traversed
forwards and $E^-(P)$ the set of edges traversed backwards.  This convention is convenient notationally, but any
initial orientation of $K_{m,n}$ will give us the same result.}. From now on, when we speak of ``oriented paths''
or ``oriented cycles'', we mean with this sign convention and some fixed traversal order.

Let $A=A_{ij}$ be a $(m\times n)$ matrix of rank $1$, and identify the entries $A_{ij}$ with the edges of
$K_{m,n}$. For an oriented cycle $C$, we define the polynomials
\begin{align*}
P_C(A) &= \prod_{e\in E^+(C)} A_e - \prod_{e\in E^-(C)} A_e,\quad\mbox{and}\\
L_C(A) &= \sum_{e\in E^+(C)} \log A_e  - \sum_{e\in E^-(C)} \log A_e,
\end{align*}
where for negative entries of $A$, we fix a branch of the complex logarithm.
\end{Def}
\begin{Thm}\label{Thm:circpoly}
Let $A=A_{ij}$ be a generic $(m\times n)$ matrix of rank $1$. Let $C\subseteq K_{m,n}$ be an oriented cycle.
Then, $P_C(A) = L_C(A) = 0$.
\end{Thm}
{\it Proof:} The determinantal ideal of rank one is a binomial ideal generated by the
$(2\times 2)$ minors of $A$ (where entries of $A$ are considered as variables).
The minor equations are exactly $P_C(A)$, where $C$ is an elementary oriented four-cycle;
if $C$ is an elementary $4$-cycle, denote its edges by $a(C)$, $b(C)$, $c(C)$, $d(C)$, with $E^+(C) =\{ a(C), d(C)\}$.
Let $\mathcal{C}$ be the collection of the elementary $4$-cycles, and define $L_\mathcal{C}(A) = \{L_C(A) : C\in\mathcal{C} \}$
and $P_\mathcal{C}(A) = \{P_C(A) : C\in\mathcal{C} \}$.   %

By sending the term $\log A_e$ to a formal variable $x_e$, we see that the free
$\mathbb{Z}$-group generated by the $L_C(A)$ is isomorphic to
$H_1(K_{m,n},\mathbb{Z})$. With this equivalence, it is straightforward
that, for any oriented
cycle $D$, $L_D(A)$ lies in the $\mathbb{Z}$-span of elements of $L_{\mathcal{C}}(A)$ and,
therefore, formally,
\[
L_D(A) = \sum_{C\in\mathcal{C}} \alpha_C\cdot L_C(A)
\]
with the $\alpha_C\in\mathbb{Z}$.  Thus $L_D(\cdot)$ vanishes when $A$ is rank one, since the r.h.s. does.
Exponentiating, we see that
\[
\left(\prod_{e\in E^+(D)} A_e \right)\left(\prod_{e\in E^-(D)} A_e \right)^{-1} =
\prod_{C\in \mathcal{C}} \left(A_{a(C)}A_{d(C)}A^{-1}_{b(C)}A^{-1}_{c(C)}\right)^{\alpha_C}
\]
If $A$ is generic and rank one, the r.h.s. evaluates to one, implying that $P_D(A)$ vanishes.\eop
\begin{Cor}\label{Cor:circpoly}
Let $A=A_{ij}$ be a $(m\times n)$ matrix of rank $1$. Let $v, w$ be two vertices in $K_{m,n}$.
Let $P,Q$ be two oriented paths in $K_{m,n}$ starting at $v$ and ending at $w$. Then, for all $A$, it holds
that $L_P(A)=L_Q(A)$.
\end{Cor}
\begin{Rem}
It is possible to prove that the set of $P_C$ forms the set of polynomials vanishing on the entries of $A$ which is minimal
with respect to certain properties. Namely, the $P_C$ form a universal Gröbner basis for the determinantal ideal of
rank $1$, which implies the converse of Theorem~\ref{Thm:circpoly}.
From this, one can deduce that the estimators presented in section~\ref{sec:est.estim} are
variance-minimal amongst all unbiased ones.
\end{Rem}

\section{A Combinatorial Algebraic Estimate for Missing Entries and Their Error}
In this section, we will construct an estimator for matrix completion which (a) is able to complete single missing
entries and (b) gives universal error estimates for that entry that are independent of the reconstruction algorithm.

\subsection{The sampling model}
\label{sec:est.sampling}
In all of the following, we will assume that the observations arise from the following sampling process:
\begin{Ass}\label{Ass:sampling}
There is an unknown fixed, rank one, matrix $A$ %
which is generic, and an $(m\times n)$ mask $M\in \{0,1\}^{m\times n}$ which is known. There is a (stochastic)
noise matrix $\calE\in\mathbb{R}^{m\times n}$ whose entries are uncorrelated and which is multiplicatively centered
with finite variance, non-zero\footnote{The zero-variance case corresponds to exact reconstruction, which is
handled already by Theorem~\ref{Thm:closure}.}
variance; i.e., $\mathbb{E}(\log \calE_{ij})= 0$ and $0 < \Var (\log \calE_{ij})<\infty$ for all
$i$ and $j$.

The observed data is the matrix $A\circ M\circ \calE = \Omega (A\circ \calE)$, where $\circ$ denotes the
Hadamard (i.e., component-wise) product. That is, the observation is a matrix with entries
$A_{ij}\cdot M_{ij}\cdot \calE_{ij}$.
\end{Ass}
The assumption of multiplicative noise is a necessary precaution in order
for the presented estimator (and in fact, any estimator) for the missing entries to have bounded variance,
as shown in Example~\ref{ex:variance} below.  This is not, in practice, a restriction since an infinitesimal
additive error $\delta A_{ij}$ on an entry of $A$ is equivalent to an
infinitesimal multiplicative error $\delta \log A_{ij} = \delta A_{ij} / A_{ij}$, and additive variances can be directly translated
into multiplicative variances if the density function for the noise is known\footnote{The multiplicative noise assumption causes
the observed entries and the true entries to have the same sign.  The change of sign can be modeled by adding another
multiplicative binary random variable in the model which takes values $\pm 1$; this adds an independent combinatorial problem
for the estimation of the sign which can be done by maximum likelihood. In order to keep the exposition short and easy,
we did not include this into the exposition.}.  The previous observation implies that the multiplicative noise model is
as powerful as any additive one that allows bounded variance estimates.
\begin{Ex}\label{ex:variance}
Consider the rank one matrix
$$A = \left(
\begin{array}{cc}
A_{11} & A_{21} \\
A_{12} & A_{22}
\end{array}\right).$$
The unique equation between the entries is $A_{11}A_{22} = A_{12}A_{21}.$ Solving for any entry will have another entry in the denominator, for example
$$A_{11} = \frac{A_{12}A_{21}}{A_{22}}.$$
Thus we get an estimator for $A_{11}$ when substituting observed and noisy entries for $A_{12},A_{21},A_{22}$.
When $A_{22}$ approaches zero, the estimation error for $A_{11}$ approaches infinity. In particular, if the density
function of the error $E_{22}$ of $A_{22}$ is too dense around the value $-A_{22}$, then the estimate for $A_{11}$
given by the equation will have unbounded variance. In such a case, one can show that no estimator for $A_{11}$ has bounded variance.
\end{Ex}

\subsection{Estimating entries and error bounds} \label{sec:est.estim}
In this section, we construct the unbiased estimator for the entries of a rank-one-matrix with minimal variance.
First, we define some notation to ease the exposition:
\begin{Not}
We will denote by $a_{ij}=\log A_{ij}$ and $\varepsilon_{ij} = \log \calE_{ij}$ the logarithmic entries and noise.
Thus, for some path $P$ in $K_{m,n}$ we obtain
$$L_P(A) = \sum_{e\in E^+(P)} a_e  - \sum_{e\in E^-(P)} a_e.$$
Denote by $b_{ij}=a_{ij}+\varepsilon_{ij}$ the logarithmic (observed) entries, and $B$ the (incomplete) matrix which
has the (observed) $b_{ij}$ as entries. Denote by $\sigma_{ij}=\Var (b_{ij}) = \Var (\varepsilon_{ij}).$
\end{Not}

The components of the estimator will be built from the $L_P$:
\begin{Lem}\label{Lem:expLPB}
Let $G=G(M)$ be the graph of the mask $M$. Let $x=(v,w)\in K_{m,n}$ be any edge with $v$ red. Let $P$ be an oriented path\footnote{If $x\in G$, then $P$ can also be the path consisting of the single edge $e$.} in $G(M)$ starting at $v$ and ending at $w$. Then,
$$L_P(B)=\sum_{e\in E^+(P)} b_e  - \sum_{e\in E^-(P)} b_e$$
is an unbiased estimator for $a_x$ with variance
$$\Var (L_P(B)) = \sum_{e\in P} \sigma_e.$$
\end{Lem}
{\it Proof:} By linearity of expectation and centeredness of $\varepsilon_{ij}$, it follows that
\begin{align*}
\mathbb{E}(L_P(B))=\sum_{e\in E^+(P)} \mathbb{E}(b_e)  - \sum_{e\in E^-(P)} \mathbb{E}(b_e),
\end{align*}
thus $L_P(B)$ is unbiased. Since the $\varepsilon_e$ are uncorrelated, the $b_e$ also are; thus, by Bienaymé's formula, we obtain
$$\Var (L_P(B)) = \sum_{e\in E^+(P)} \Var(b_e)  + \sum_{e\in E^-(P)} \Var(b_e),$$
and the statement follows from the definition of $\sigma_e.$\\

In the following, we will consider the following parametric estimator as a candidate for estimating $a_e$:
\begin{Not}\label{Not:calPXalpha}
Fix an edge $x=(v,w)\in K_{m,n}$. Let $\calP$ be a basis
for the set of all oriented paths starting at $v$ and ending at $w$
\footnote{This is the set of words equal to the formal generators $x_{(v,w)}$ in the free abelian
group generated by the $x_e$, subject to the relations $L_C = 0$ for all cycles $C$ in $G\cup\{(v,w)\}$.
Independence can be taken as linear independence of the coefficient vectors of the
$L_C$.},
and denote $\#\calP$ by $p$.
For $\alpha\in\mathbb{R}^p$, set
\[
X(\alpha)= \sum_{P\in\calP} \alpha_P L_P(B).
\]

Furthermore, we will denote by $\mathbbm {1}$ the $n$-vector of ones.
\end{Not}

The following Lemma follows immediately from Lemma~\ref{Lem:expLPB} and Theorem~\ref{Thm:circpoly}:
\begin{Lem}
$\mathbb{E}(X(\alpha)) = \mathbbm {1}^\top \alpha \cdot b_x;$ in particular, $X(\alpha)$ is an unbiased estimator
for $b_x$ if and only if $\mathbbm {1}^\top \alpha=1$.
\end{Lem}

We will now show that minimizing the variance of $X(\alpha)$ can be formulated as a quadratic program with
coefficients entirely determined by $a_x$, the measurements $b_e$ and the graph $G(M)$. In particular, we will
expose an explicit formula for the $\alpha$ minimizing the variance.  Before stating the theorem, we define a suitable kernel:
\begin{Def}
Let $e\in K_{m,n}$ be an edge. For an edge $e$ and a path $P$, set $c_{e,P}=\pm1$ if $e\in E^\pm (P)$ otherwise $c_{e,P}=0.$
Let $P,Q\in\calP$ be any fixed oriented paths. Define the (weighted) {\it path kernel}
$k:\calP\times \calP\rightarrow \mathbb{R}$ by
$$k(P,Q)=\sum_{e\in K_{m,n}} c_{e,P}\cdot c_{e,Q}\cdot \sigma_e.$$
\end{Def}
Under our assumption that $\Var(b_e)>0$ for all $e\in K_{m,n}$, the path kernel is positive definite, since it
is a sum of $p$ independent positive semi-definite functions; in particular, its kernel matrix has full rank.
Here is the variance-minimizing unbiased estimator:
\begin{Prop}\label{Prop:Varest}
Let $x=(s,t)$ be a pair of vertices, and $\calP$ a basis for the $s$--$t$ path space in $G$ with $p$ elements.  Let
$\Sigma$ be the $p\times p$ kernel matrix of the path kernel with respect to the basis $\calP$.
For any $\alpha\in\mathbb{R}^p$,
\[
\Var(X(\alpha))=\alpha^\top \Sigma\alpha.
\]
Moreover, under the condition $\mathbbm {1}^\top\alpha =1$, the variance $\Var(X(\alpha))$ is minimized
by
\[
\alpha = \left(\Sigma^{-1}\mathbbm {1}\right)\left(\mathbbm {1}^\top \Sigma^{-1}\mathbbm {1}\right)^{-1}
\]
\end{Prop}
{\it Proof:} By inserting definitions, we obtain
\begin{align*}
X(\alpha)&= \sum_{P\in\calP} \alpha_P L_P(B)\\
&=\sum_{P\in\calP} \alpha_P \sum_{e\in K_{m,n}} c_{e,P} b_e .
\end{align*}
Writing $b=(b_e)\in\mathbb{R}^{mn}$ as vectors, and $C=(c_{e,P})\in\mathbb{R}^{p\times mn}$ as matrices, we obtain
$$X(\alpha) = b^\top C \alpha.$$
By using that $\Var(\lambda\cdot )=\lambda^2\Var(\cdot)$ for any scalar $\lambda$, and independence of the
$b_e$, an elementary calculation yields
$$\Var(X(\alpha)) = \alpha^\top \Sigma \alpha$$
In order to determine the minimum of the variance in $\alpha$, consider the Lagrangian
$$L(\alpha,\lambda)= \alpha^\top \Sigma\alpha +  \lambda \left(1- \sum_{P\in\calP}\alpha_P\right),$$
where the slack term models the condition $\ell (\alpha)=1$. An elementary calculation yields
\begin{align*}
\frac{\partial L}{\partial\alpha}& = 2\Sigma\alpha - \lambda \mathbbm {1}
\end{align*}
where $\mathbbm {1}$ is the vector of ones. Due to positive definiteness of $\Sigma$ the function $\Var(X(\alpha))$ is convex,
thus $\alpha = \Sigma^{-1}\mathbbm {1}/\mathbbm {1}^\top \Sigma^{-1}\mathbbm {1}$ will be the unique $\alpha$ minimizing
the variance while satisfying $\mathbbm {1}^\top\alpha =1$.\eop
\begin{Rem}
The above setup works in wider generality: (i) if $\Var(b_e)=0$ is allowed and there is an $s$--$t$ path of
all zero variance edges, the path kernel becomes positive semi-definite; (ii) similarly if $\mathcal{P}$
is replaced with any set of paths at all, the same may occur.  In both cases, we may replace $\Sigma^{-1}$
with the Moore-Penrose pseudo-inverse and the proposition still holds: (i) reduces to the exact
reconstruction case of Theorem~\ref{Thm:closure}; (ii) produces the optimal estimator with respect to $\mathcal{P}$,
which is optimal provided that $\mathcal{P}$ is spanning, and adding paths to $\mathcal{P}$ does not make the estimate
worse.
\end{Rem}

\subsection{Rank $2$ and higher}
An estimator for rank $2$ and higher, together with a variance analysis, can be constructed similarly once all polynomials known which relate the entries under each other. The main difficulty lies in the fact that these polynomials are not parameterized by cycles anymore, but specific subgraphs of $G(M)$, see~\citep[Section 2.5]{KTTU12}. Were these polynomials known, an estimator similar to $X(\alpha)$ as in Notation~\ref{Not:calPXalpha} could be constructed, and a subsequent variance (resp.~perturbation) analysis performed.

\subsection{The algorithms}
In this section, we describe the two main algorithms which calculate the variance-minimizing estimate $\widehat{A}_{ij}$ for any
fixed  entry $A_{ij}$ of an $(m\times n)$ matrix $A$, which is observed with noise, and the variance bound for the
estimate $\widehat{A}_{ij}$. It is important to note that $A_{ij}$ does not necessarily need to be an entry which is
missing in the observation, it can also be any entry which has been observed. In the latter case, Algorithm~\ref{Alg:Xalpha}
will give an improved estimate of the observed entry, and Algorithm~\ref{Alg:Var} will give the trustworthiness
bound on this estimate.

Since the the path matrix $C$, the path kernel matrix $\Sigma$, and the optimal $\alpha$ is required for both,
we first describe Algorithm~\ref{Alg:alpha} which determines those.
\begin{algorithm}[h]
\caption[Calculates path kernel $\Sigma$ and $\alpha$]{Calculates path kernel $\Sigma$ and $\alpha$.\\
\textit{Input:} index $(i,j),$ an $(m\times n)$ mask $M$, variances $\sigma$.\\
\textit{Output:} path matrix $C$, path kernel $\Sigma$ and minimizer $\alpha$. \label{Alg:alpha}}
\begin{algorithmic}[1]
\item[1:] Find a linearly independent set of paths $\calP$ in the graph $G(M)$, starting from $i$ and ending at $j$.\label{Alg:alpha.step1}
\item[2:] Determine the matrix $C=(c_{e,P})$ with $e\in G(M), P\in \calP;$ set $c_{e,P} = \pm 1$ if $e\in E^\pm (P)$, otherwise $c_{e,P}=0.$
\item[3:] Define a diagonal matrix $S = \diag(\sigma),$ with $S_{ee}=\sigma_e$ for $e\in G(M).$
\item[4:] Compute the kernel matrix $\Sigma = C^\top S C.$
\item[5:] Calculate $\alpha = \Sigma^{-1}\mathbbm {1}/\|\Sigma^{-1}\mathbbm {1}\|_1.$
\item[6:] Output $C,\Sigma$ and $\alpha$.
\end{algorithmic}
\end{algorithm}
The steps of the algorithm
follow the exposition in section~\ref{sec:est.estim}, correctness follows from the statements presented there.
The only
task in Algorithm~\ref{Alg:alpha} that isn't straightforward is the computation of a linearly independent set of paths
in step~\ref{Alg:alpha.step1}.  We can do this time linear in the number of observed entries in the mask $M$
with the following method.  To keep the notational manageable,
we will conflate formal sums of the $x_e$, cycles in $H_1(G,\mathbb{Z})$ and their representations as vectors
in $\mathbb{R}^{mn}$, since there is no chance of confusion.
\begin{algorithm}[h]
\caption[Calculates a basis $\calP$ of the path space]{Calculates a basis $\calP$ of the path space.\\
\textit{Input:} index $(i,j),$ an $(m\times n)$ mask $M$.\\
\textit{Output:} a basis $\calP$ for the space of oriented $i$--$j$ paths. \label{Alg:calP}}
\begin{algorithmic}[1]
\item[1:] If $(i,j)$ is not an edge of $M$, and $i$ and $j$ are in different connected components,
then $\calP$ is empty.
Output $\emptyset$.
\item[2:] Otherwise, if $(i,j)$ is not an edge, of $M$, add a ``dummy'' copy.
\item[3:] Compute a spanning forest $F$ of $M$ that does not contain $(i,j)$, if possible.
\item[4:] For each edge $e\in M\setminus F$, compute the fundamental cycle $C_e$ of
$e$ in $F$.
\item[5:] If $(i,j)$ is an edge in $M$, output $\{-x_{(i,j)}\} \cup \{C_e - x_{(i,j)} : e\in M\setminus F\}$.
\item[6:] Otherwise, let $P_{(i,j)} = C_{(i,j)} - x_{(i,j)}$.  Output $\{C_e - P_{(i,j)} : e\in M\setminus (F\cup \{(i,j)\})\}$.
\end{algorithmic}
\end{algorithm}
We prove the correctness of Algorithm~\ref{Alg:calP}.

Algorithms~\ref{Alg:Xalpha} and~\ref{Alg:Var} then can make use of the calculated $C,\alpha,\Sigma$ to determine an
estimate for any entry $A_{ij}$ and its minimum variance bound. The algorithms follow the exposition in
Section~\ref{sec:est.estim}, from where correctness follows; Algorithm~\ref{Alg:Xalpha} additionally
provides treatment for the sign of the entries.
\begin{algorithm}[h]
\caption[Estimates the entry $a_{ij}$.]{\label{Alg:Xalpha} Estimates the entry $a_{ij}$.\\
\textit{Input:} index $(i,j),$ an $(m\times n)$ mask $M$, log-variances $\sigma$, the partially observed and noisy matrix $B$.\\
\textit{Output:} The variance-minimizing estimate for $A_{ij}$. }
\begin{algorithmic}[1]

\item[1:] Calculate $C$ and $\alpha$ with Algorithm~\ref{Alg:alpha}.
\item[2:] Store $B$ as a vector $b=(\log |B_e|)$ and a sign vector $s=(\sgn B_e)$ with $e\in G(M)$.
\item[3:] Calculate $\widehat{A}_{ij} = \pm \exp \left(b^\top C \alpha\right).$ The sign is $+$ if each column of $s^\top |C|$ ($|.|$ component-wise)  contains an odd number of entries $-1$, else $-$.
\item[4:] Return $\widehat{A}_{ij}.$
\end{algorithmic}
\end{algorithm}

\begin{algorithm}[h]
\caption[Determines the variance of the entry $\log(A_{ij})$]{\label{Alg:Var} Determines the variance of the entry $\log(A_{ij})$.\\
\textit{Input:} index $(i,j),$ an $(m\times n)$ mask $M$, log-variances $\sigma$.\\
\textit{Output:} The variance lower bound for $\log (A_{ij})$. }
\begin{algorithmic}[1]

\item[1:] Calculate $\Sigma$ and $\alpha$ with Algorithm~\ref{Alg:alpha}.
\item[2:] Return $\alpha^\top \Sigma\alpha$.
\end{algorithmic}
\end{algorithm}

Note that even if observations are not available, Algorithm~\ref{Alg:Var} can be used to obtain the variance bound.
The variance bound is relative, due to its multiplicativity, and can be used to approximate absolute bounds when
any reconstruction estimate $\widehat{A}_{ij}$ is available - which does not necessarily need to be the one from
Algorithm~\ref{Alg:Xalpha}, but can be the estimation result of any reconstruction. Namely, if $\widehat{\sigma}_{ij}$
is the estimated variance of the log, we obtain an upper confidence bound (resp.~deviation) bound
$\widehat{A}_{ij}\cdot \exp{\left(\sqrt{\widehat{\sigma}_{ij}}\right)}$ for $\widehat{A}_{ij},$ and a lower
confidence bound (resp.~deviation) bound $\widehat{A}_{ij}\cdot \exp{\left(-\sqrt{\widehat{\sigma}_{ij}}\right)}$,
corresponding to the log-confidence $\log \widehat{A}_{ij}\pm\sqrt{\widehat{\sigma}_{ij}}.$ Also note that if $A_{ij}$
is not reconstructible from the mask $M$ (i.e., if the edge $(i,j)$ is not in the transitive
closure of $G(M)$, see Theorem~\ref{Thm:closure}), then the deviation bounds will be infinite.

\section{Experiments}\label{sec:exp}

\subsection{Universal error estimates}
\begin{figure*}[ht]
\begin{center}
\subfigure{%
\includegraphics[height=0.3\textwidth]{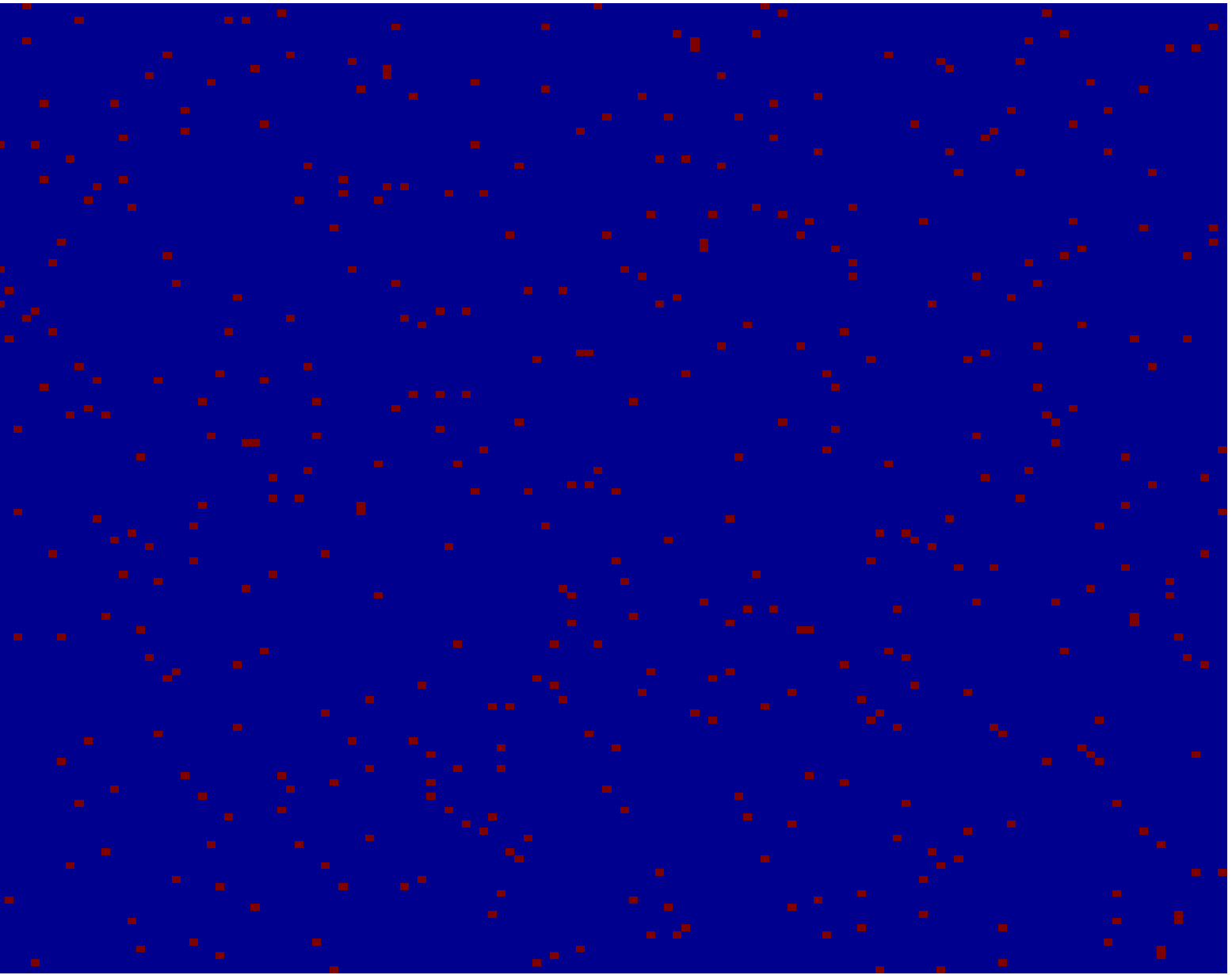}
}
\subfigure{%
\includegraphics[height=0.3\textwidth]{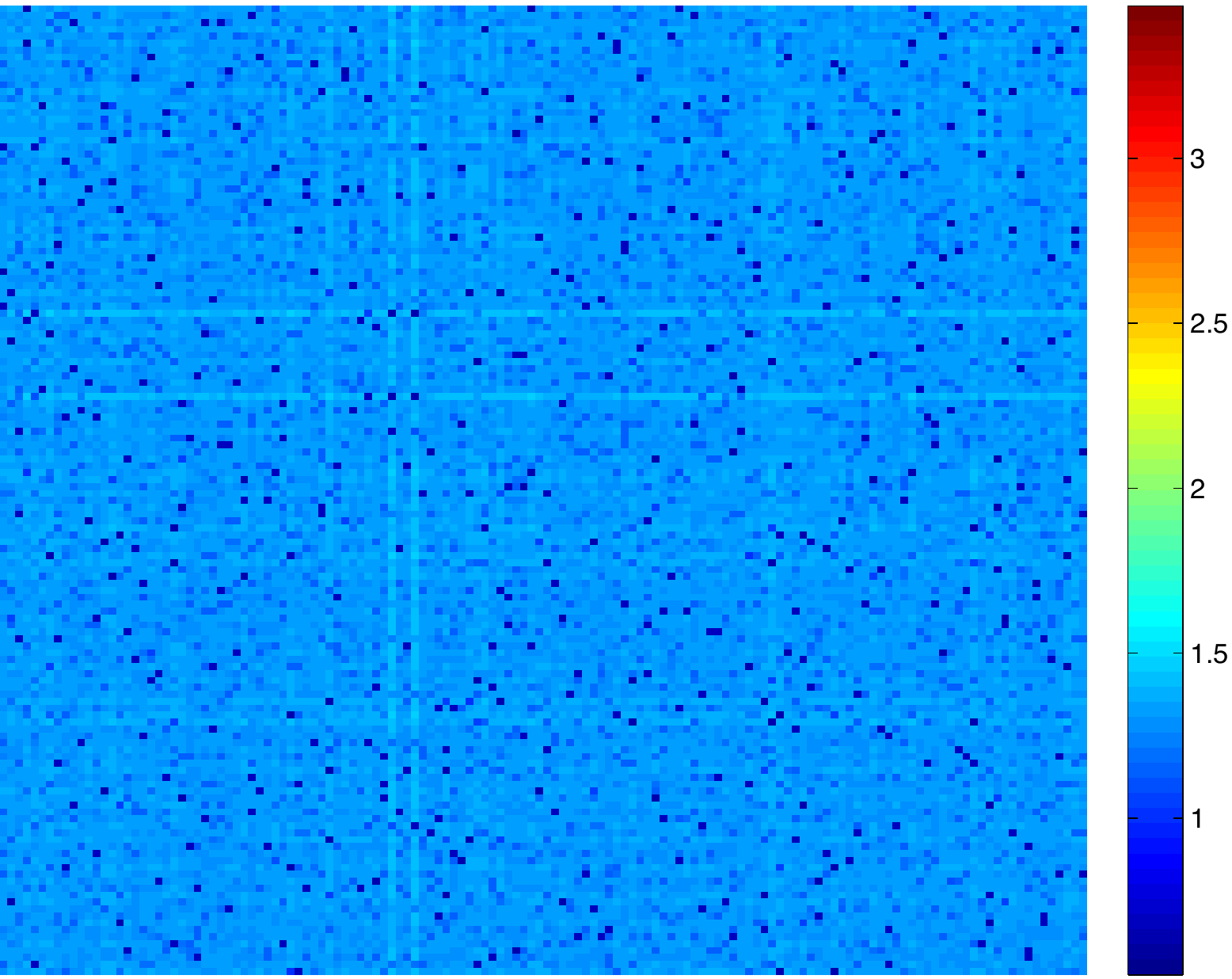}
}\\
\subfigure{%
\includegraphics[height=0.3\textwidth]{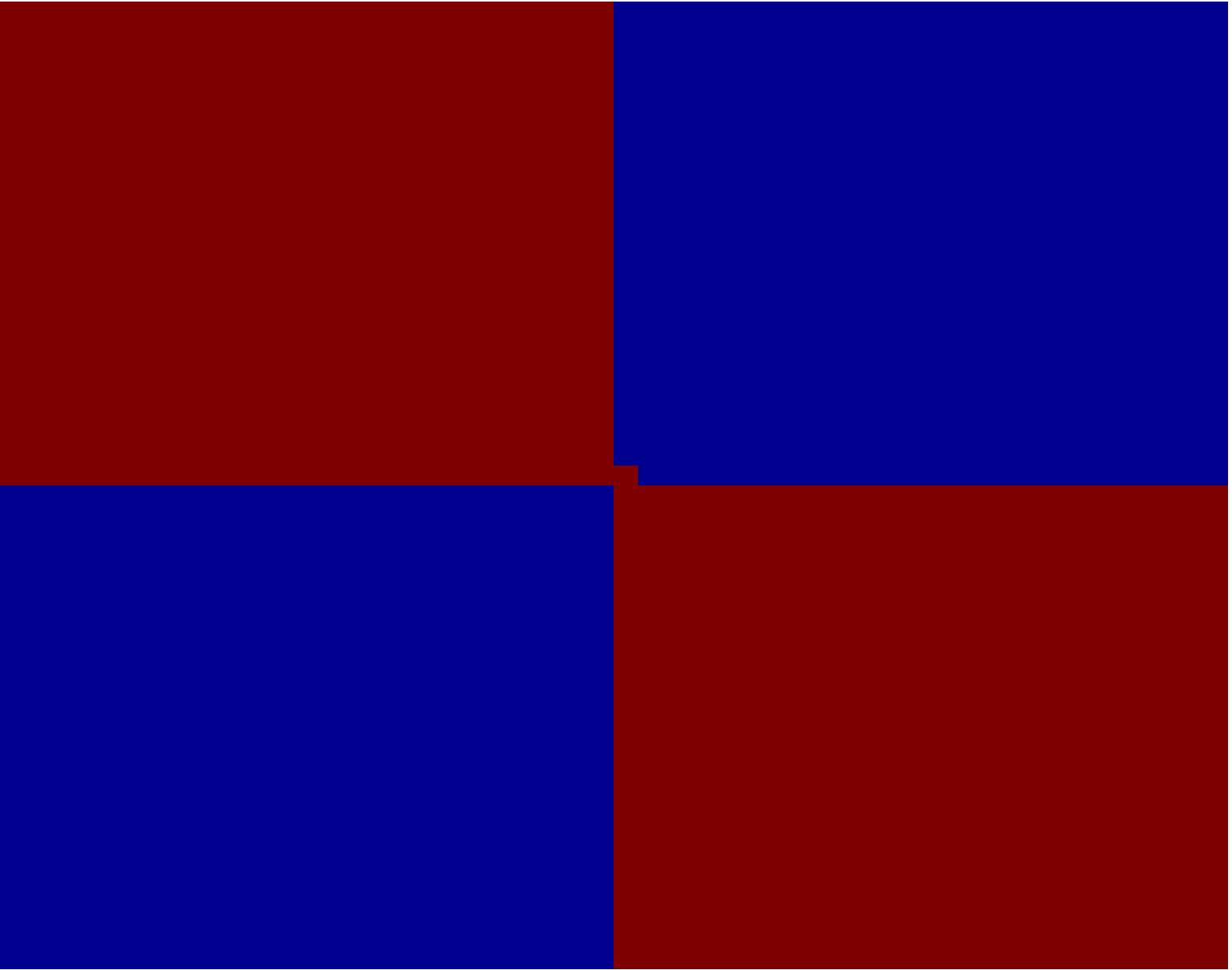}
}
\subfigure{%
\includegraphics[height=0.3\textwidth]{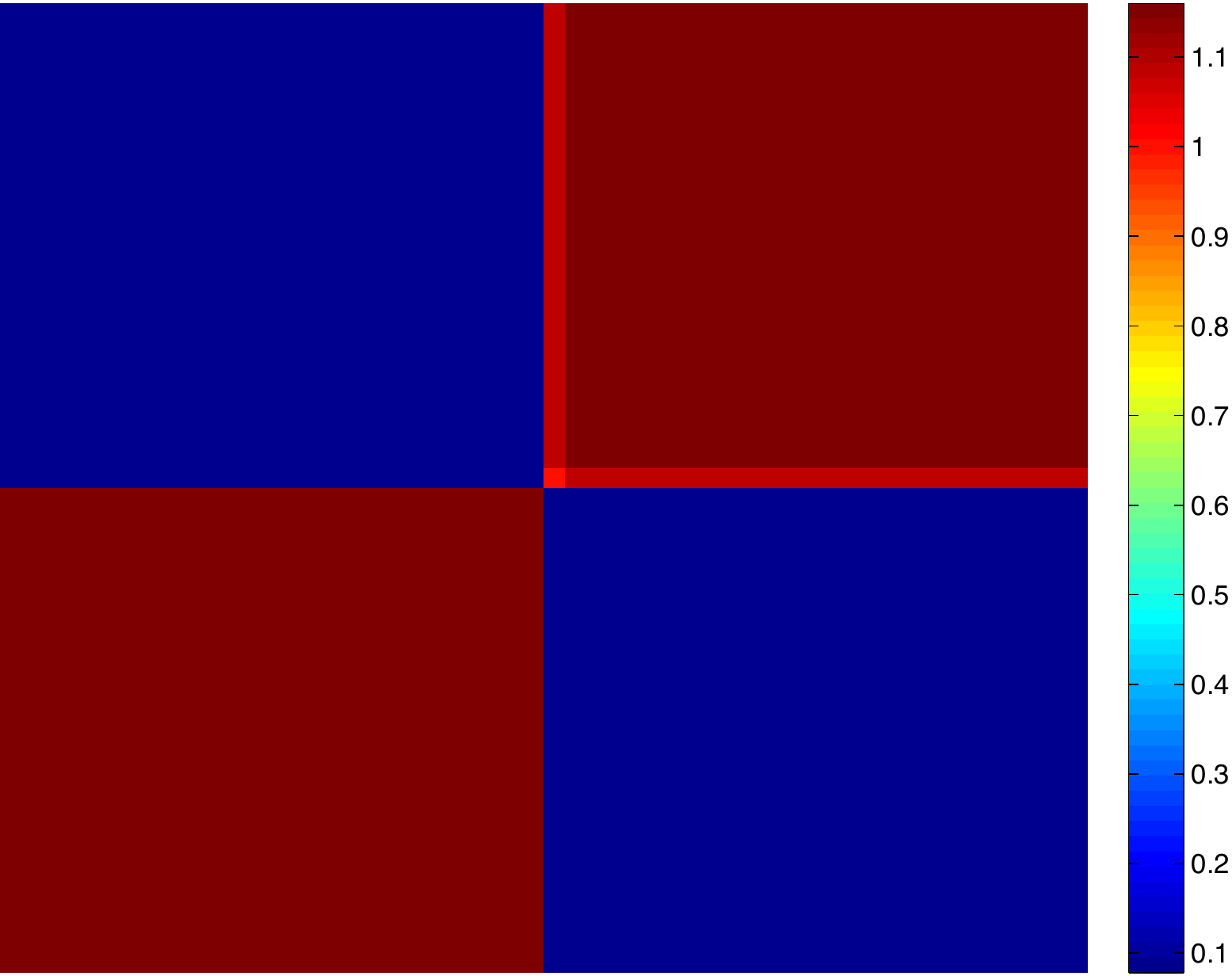}
}\\
\subfigure{%
\includegraphics[height=0.3\textwidth]{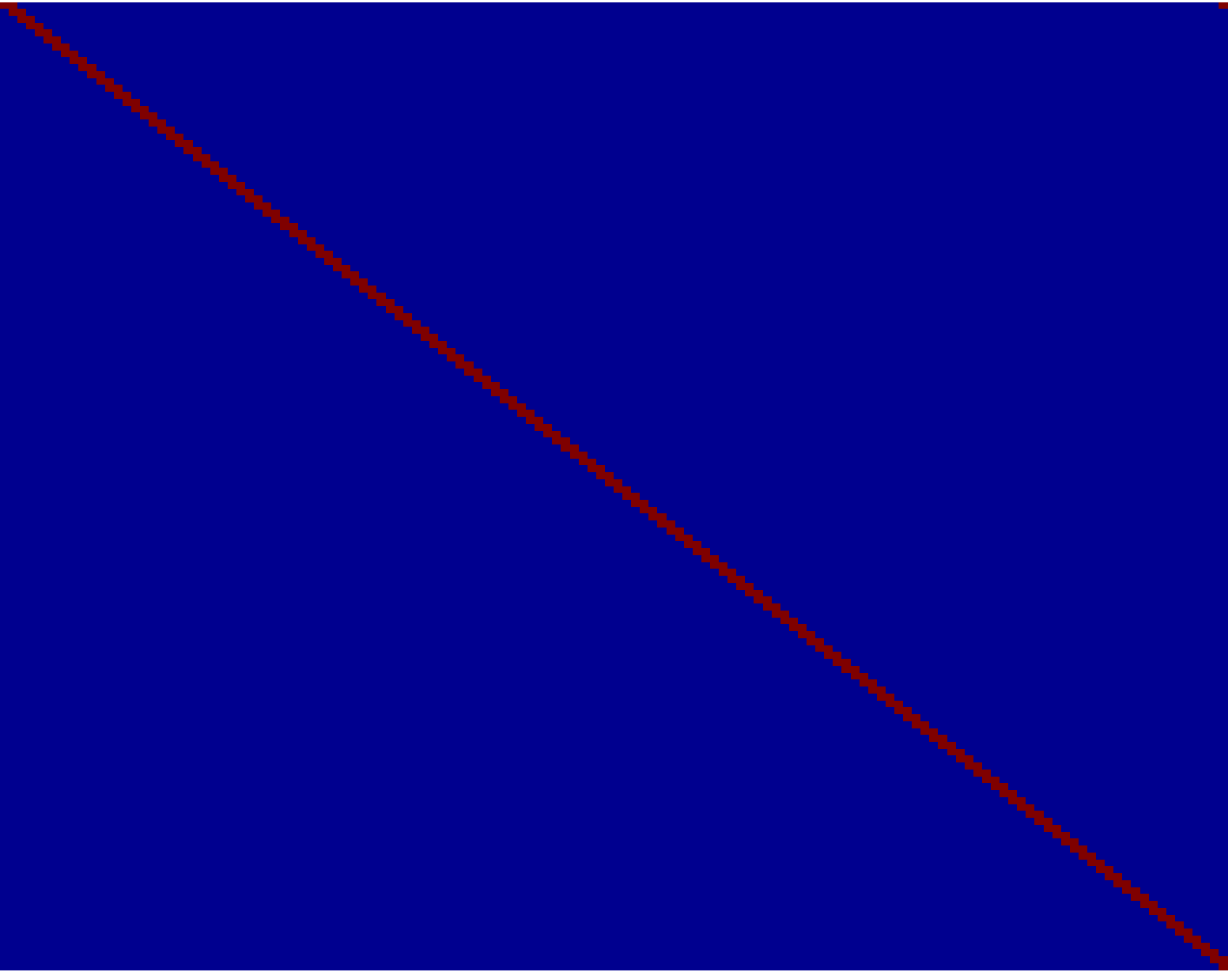}
}
\subfigure{%
\includegraphics[height=0.3\textwidth]{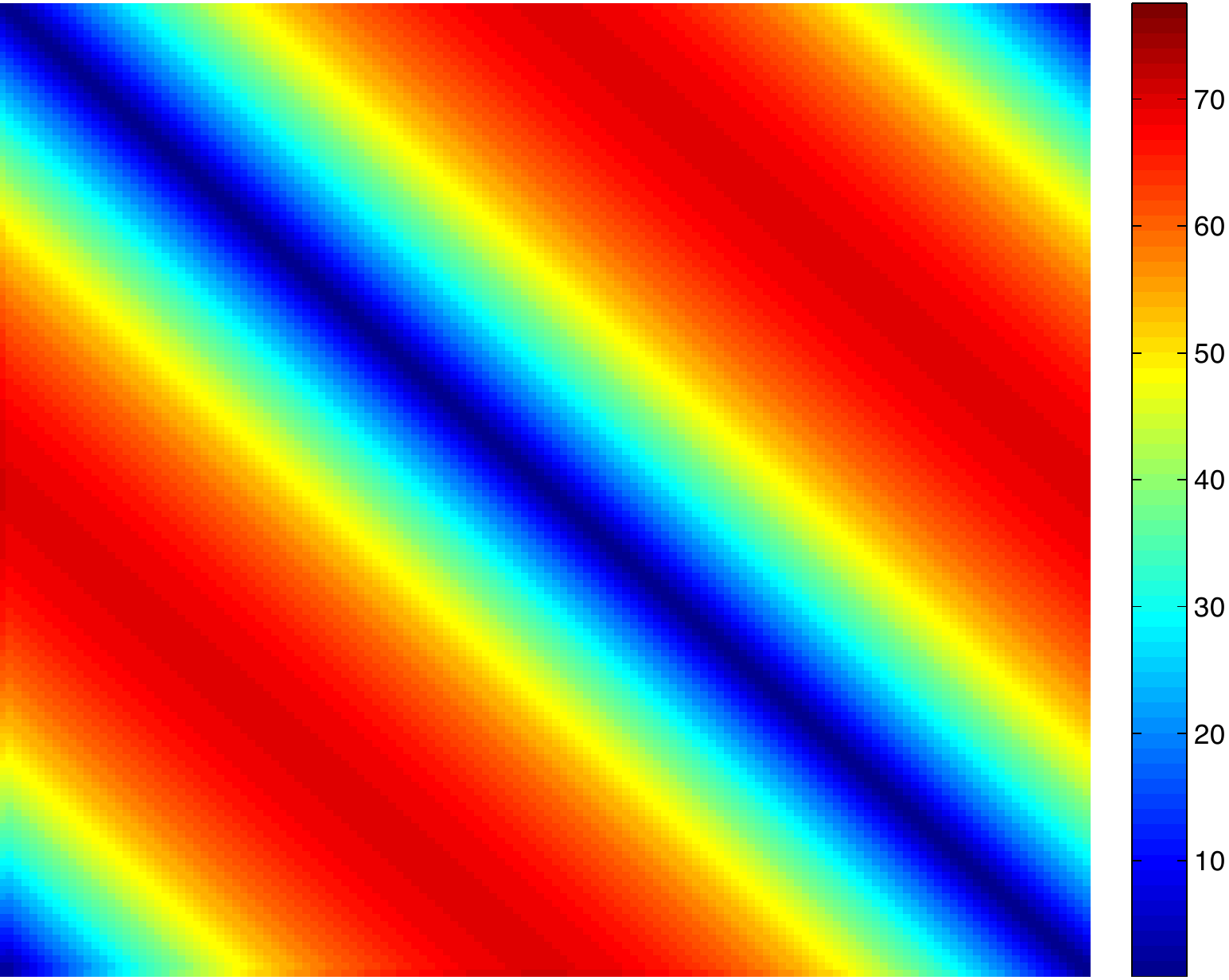}
}
\end{center}
\caption{%
\label{fig:heatmaps}
The figure shows three pairs of masks and predicted variances. A pair consists of two adjacent squares. The left half is the mask which is depicted by red/blue heatmap with red entries known and blue unknown. The right half is a multicolor heatmap with color scale, showing the predicted variance of the completion. Variances were calculated by our implementation of Algorithm~\ref{Alg:Var}.
}
\end{figure*}

For three different masks, we calculated the predicted minimum variance for each entry of the mask.
The multiplicative noise was assumed to be $\sigma_e=1$ for each entry. Figure~\ref{fig:heatmaps} shows the predicted
a-priori minimum variances for each of the masks. Notice how the structure of the mask affects the expected error; known
entries generally have least variance, while it is interesting to note that in general it is less than the starting
variance of $1$. I.e., tracking back through the paths can be successfully used even to denoise known entries.
The particular structure of the mask is mirrored in the pattern of the predicted errors; a diffuse mask gives a
similar error on each missing entry, while the more structured masks have structured error which is determined by
combinatorial properties of the completion graph and the paths therein.

\begin{figure*}[ht]
\begin{center}
\subfigure[mean squared errors]{%
\label{fig:3algs.noisemse}
\includegraphics[width=0.4\textwidth]{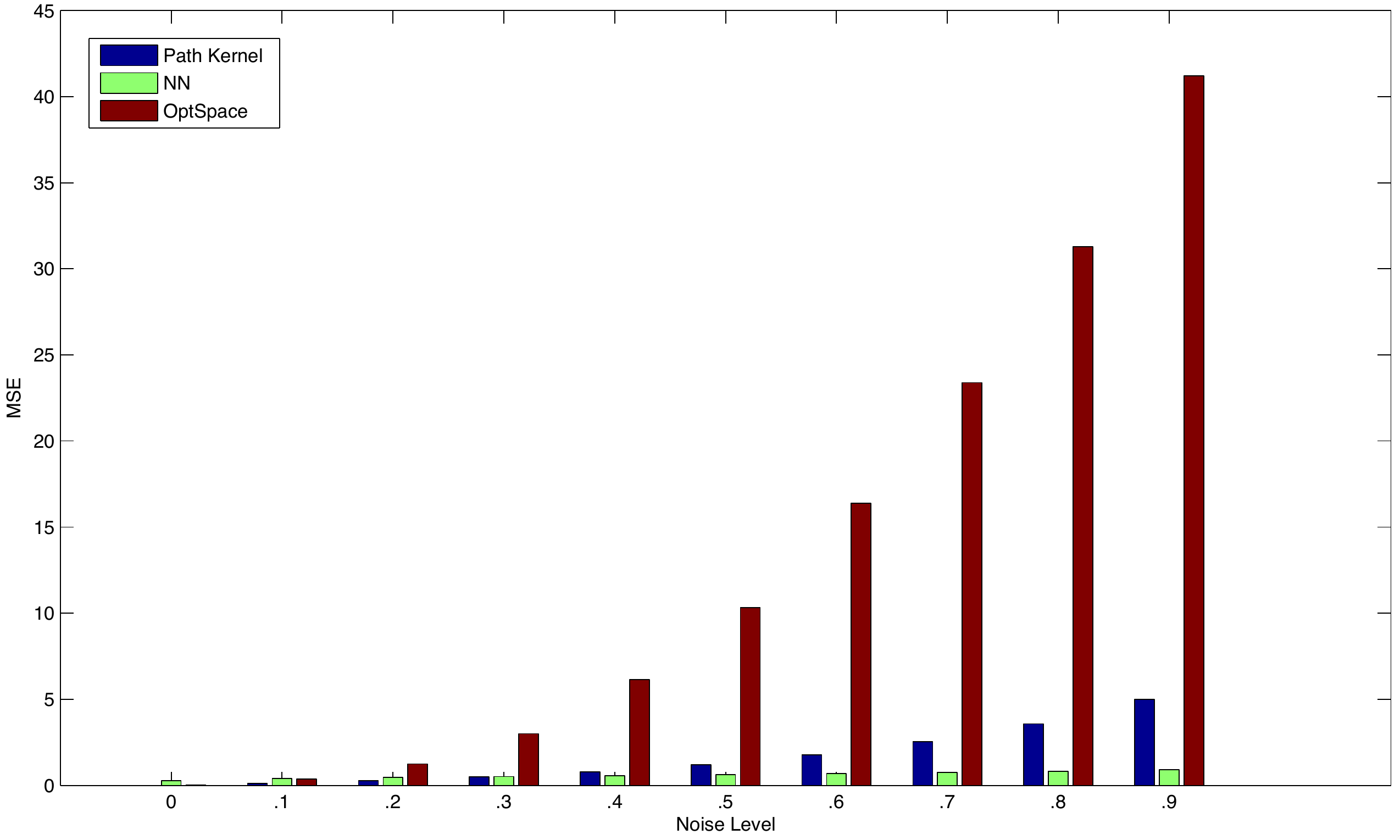}
}
\subfigure[error vs.~predicted variance]{%
\label{fig:3algs.linearity}
\includegraphics[width=0.4\textwidth]{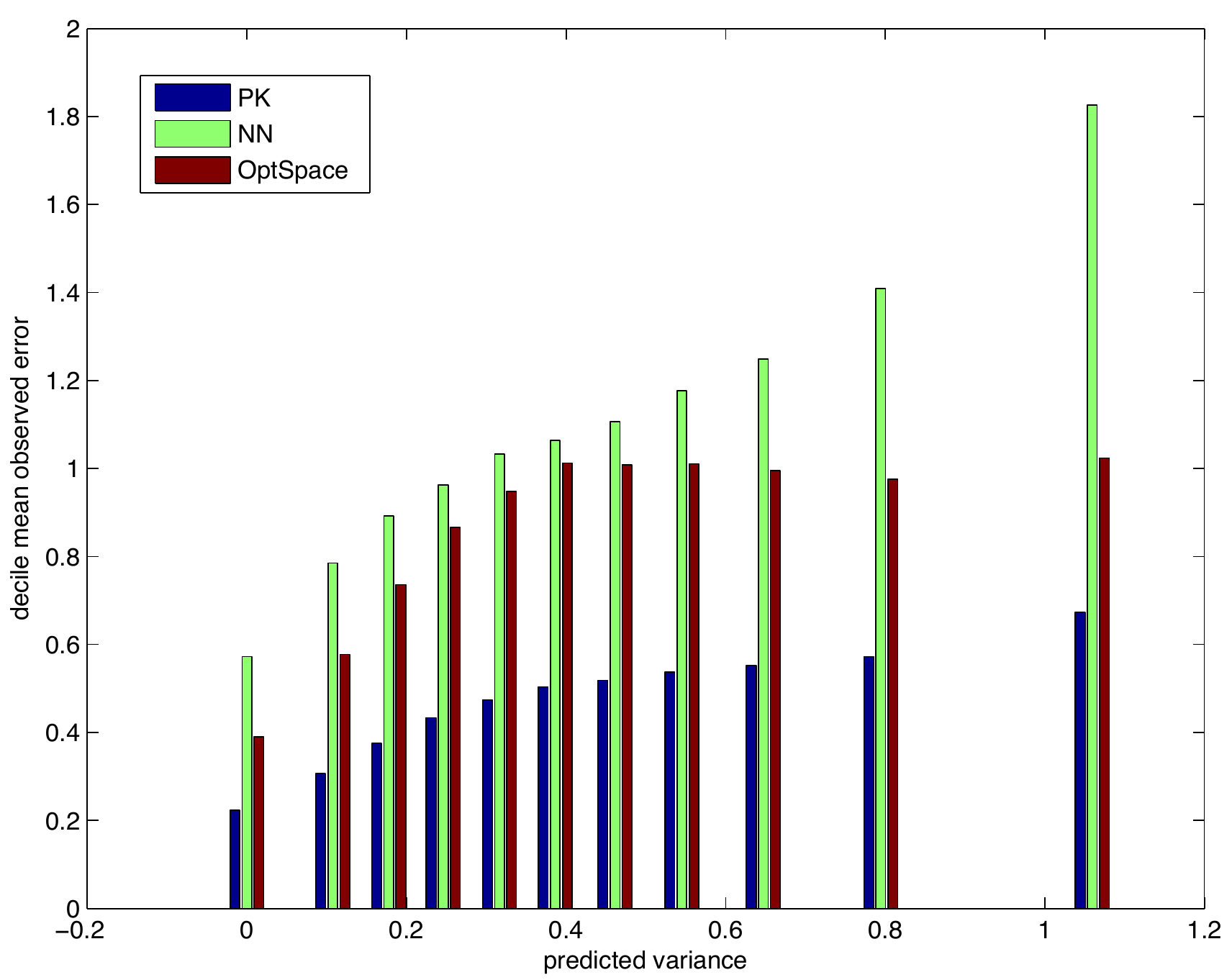}
}

\end{center}
\caption{%
For $10$ randomly chosen masks and $50\times 50$ true matrix, matrix completions were performed with Nuclear Norm (green), OptSpace (red), and Algorithm~\ref{Alg:Xalpha} (blue) under multiplicative noise with variance increasing in increments of $0.1$. For each completed entry, minimum variances were predicted by Algorithm~\ref{Alg:Var}. \ref{fig:3algs.noisemse} shows the mean squared error of the three algorithms for each noise level, coded by the algorithms' respective colors. \ref{fig:3algs.linearity} shows a bin-plot of errors (y-axis) versus predicted variances (x-axis) for each of the three algorithms: for each completed entry, a pair (predicted error, true error) was calculated, predicted error being the predicted variance, and the actual prediction error measured as $\log\abs$ of prediction minus $\log\abs$ of true entry. Then, the points were binned into $11$ bins with equal numbers of points. The figure shows the mean of the errors (second coordinate) of the value pairs with predicted variance (first coordinate) in each of the bins, the color
corresponds to the particular algorithm; each group of bars is centered on the minimum value of the associated bin.
}%
\end{figure*}

\subsection{Influence of noise level}\label{Sec:noise}
We generated $10$ random mask of size $50\times 50$ with $200$ entries sampled uniformly and a
random $(50\times 50)$ matrix of rank one.  The multiplicative noise was chosen entry-wise independent, with variance $\sigma_{i} = (i-1)/10$ for each entry. Figure \ref{fig:3algs.noisemse}
compares the Mean Squared Error (MSE) for three algorithms: Nuclear Norm (using the implementation \citet{THK10}),
OptSpace \citep{KMO10}, and Algorithm~\ref{Alg:Xalpha}. It can be seen that on this particular mask, Algorithm~\ref{Alg:Xalpha} is competitive with the other methods and even outperforms them for low noise.

\subsection{Prediction of estimation errors}
The data are the same as in Section~\ref{Sec:noise}, as are the compared algorithm. Figure \ref{fig:3algs.linearity} compares the error of each of the methods with the variance predicted by Algorithm~\ref{Alg:Var} each time the noise level changed. The figure shows that for any of the algorithms, the mean of the actual error increases with the predicted error, showing that the error estimate is useful for a-priori prediction of the actual error  - independently of the particular algorithm. Note that by construction of the data this statement holds in particular for entry-wise predictions. Furthermore, in quantitative comparison Algorithm~\ref{Alg:Var} also outperforms the other two in each of the bins.

\section{Conclusion}
In this paper, we have introduced an algebraic combinatorics based method for reconstructing and denoising single entries of an incomplete and noisy matrix, and for calculating confidence bounds of single entry estimations for arbitrary algorithms. We have evaluated these methods against state-of-the art matrix completion methods. The results of section~\ref{sec:exp} show that our reconstruction method is competitive and that - for the first time - our variance estimate provides a reliable prediction of the error on each single entry which is an a-priori estimate, i.e., depending only on the noise model and the position of the known entries.
Furthermore, our method allows to obtain the reconstruction and the error estimate for a single entry which existing methods are not capable of, possibly using only a small subset of neighboring entries - a property which makes our method unique and particularly attractive for application to large scale data. We thus argue that the investigation of the algebraic combinatorial properties of matrix completion, in particular in rank $2$ and higher where these are not yet completely understood, is crucial for the future understanding and practical treatment of big data.

\bibliographystyle{plainnat}

\newpage
\appendix
\section{Correctness of Algorithm~\ref{Alg:calP}}
We adopt the conventions of Section~\ref{sec:theory}, so that $G$ is a
bipartite graph with $m$ blue vertices, $n$ red ones, and $e$ edges
oriented from blue to red.
Recall the isomorphism, observed in the proof of Theorem~\ref{Thm:circpoly} of the
$\Z$-group of the polynomials $L_C(\cdot)$ and the oriented cycle space
$H_1(G,\mathbb{Z})$.

Define $\beta_1(G) = e - n - m + c$ (the first Betti number of the graph).
Some standard facts are that: (i)
the rank of $H_1(G,\mathbb{Z})$ is $\beta_1(G)$; (ii) we can obtain a basis
for $H_1(G,\mathbb{Z})$ consisting only of simple cycles
by picking any spanning forest $F$ of $G$ and then using as basis elements the
fundamental cycles $C_e$ of the edges $e\in E\setminus F$.
This justifies step 4.

Let $(i,j)$ be an edge of $G$.  Define an $i$--$j$ to be the
set of subgraphs such that, for generic rank one $A$, $L_P(A) = -x_{(i,j)}$.
By Theorem~\ref{Thm:circpoly}, we can write these as $\Z$-linear
combinations of $x_{(i,j)}$ and oriented cycles.
From this, we see that the rank of the path space is $\beta_1(G)+1$
and the graph theoretic identification of elements in the path space with subgraphs that have
even degree at every vertex except $i$ and $j$.  Thus, if $(i,j)$ is an edge of $G$, step
5 is justified, completing the proof of correctness in this case.

If $(i,j)$ was not an edge, step 1 guarantees that the dummy copy of $(i,j)$ that we added
is not in the spanning tree computed in step 3.  Thus, the element $P_{(i,j)} = C_{(i,j)} - x_{(i,j)}$
computed in step 6 is a simple path from $i$ to $j$.  The collection of elements generated in step
6 is independent by the same fact in $H_1(G\cup\{(i,j)\},\mathbb{Z})$ and has rank $\beta_1(G) + 1$
and does not put a positive coefficient on the dummy generator $x_{(i,j)}$. \eop
\end{document}